\documentclass[10pt,twocolumn,letterpaper]{article}
\newif\ifcomments
\commentsfalse
\commentstrue

\usepackage{cvpr}
\usepackage{times}
\usepackage{epsfig}
\usepackage{graphicx}
\usepackage{amsmath}
\usepackage{amssymb}
\usepackage{wrapfig}
\usepackage{multirow}
\usepackage{placeins}
\usepackage{subcaption}
\usepackage{caption}
\usepackage[T1]{fontenc}
\usepackage[font=small,labelfont=bf,tableposition=top]{caption}

\DeclareCaptionLabelFormat{andtable}{#1~#2  \&  \tablename~\thetable}

\usepackage[pagebackref=true,breaklinks=true,letterpaper=true,colorlinks,bookmarks=false]{hyperref}

\cvprfinalcopy 

\ifcvprfinal\fi
\begin{document}

\ifcomments
  \newcommand{\comments}[1]{#1}
\else
  \newcommand{\comments}[1]{}
\fi

\newcommand{\name}{Deep Feature Interpolation}
\newcommand{\nameshort}{DFI}

\newcommand{\kqw}[1]{\comments{\textcolor{blue}{[Kilian: #1]}}}

\title{{Addressing Vulnerabilities in AI-Image Detection: Challenges and Proposed Solutions}}

\author{{Justin Jiang}\\
{Independent Researcher}\\
{\tt\small {justin.jiang.co@gmail.com}}
}

\newcommand{\methodname}{dense convolutional  network}
\newcommand{\methodnamecap}{Dense Convolutional Network}
\newcommand{\methodnameshort}{DenseNet}
\newcommand{\methodnameshorts}{DenseNets}
\newcommand{\methodblock}{dense block}
\newcommand{\methodblockcap}{Dense Block}

\newcommand{\regmethodname}{feature drop}
\newcommand{\regmethodnamecap}{Feature Drop}

\newcommand{\stepsizename}{growth rate}

\newcommand{\conv}[1]{$\left[\begin{array}{ll} \text{1}\times \text{1} \text{ conv}\\ \text{3}\times \text{3} \text{ conv} \end{array}\right] \times \text{#1}$}

\newcommand{\cross}[1]{#1 $\times$ #1}

\newcommand{\feati}{x_i}
\newcommand{\clsfeati}{y_i}
\newcommand{\featk}{x_k}
\newcommand{\clsfeatk}{y_k}
\newcommand{\loss}{L}
\newcommand{\featL}{x_L}
\newcommand{\clsfeat}{y}
\newcommand{\anyxs}{\ensuremath{\mathbf{x}}}
\newcommand{\anyys}{\ensuremath{\mathbf{y}}}

\newcommand{\bx}{\ensuremath{\mathbf{x}}}

\newcommand{\sourcexs}{\ensuremath{\mathbf{x^\mathcal{S}}}}
\newcommand{\sourceys}{\ensuremath{\mathbf{y^\mathcal{S}}}}

\newcommand{\targetxs}{\ensuremath{\mathbf{x^\mathcal{T}}}}
\newcommand{\targetys}{\ensuremath{\mathbf{y^\mathcal{T}}}}
\newcommand{\pseudotargetys}{\ensuremath{\mathbf{\hat{y}^\mathcal{T}}}}

\maketitle

\begin{abstract}
The rise of advanced AI models like Generative Adversarial Networks (GANs) and diffusion models such as Stable Diffusion has made the creation of highly realistic images accessible, posing risks of misuse in misinformation and manipulation. This study evaluates the effectiveness of convolutional neural networks (CNNs), as well as DenseNet architectures, for detecting AI-generated images. Using variations of the CIFAKE dataset, including images generated by different versions of Stable Diffusion, we analyze the impact of updates and modifications such as Gaussian blurring, prompt text changes, and Low-Rank Adaptation (LoRA) on detection accuracy. The findings highlight vulnerabilities in current detection methods and propose strategies to enhance the robustness and reliability of AI-image detection systems.

\end{abstract}


\section{Introduction}

The rapid advancement of artificial intelligence (AI) has led to significant improvements in image generation techniques, resulting in AI-generated images that are increasingly indistinguishable from real photographs~\cite{nightingale2022aisynthesized}. Models such as Generative Adversarial Networks (GANs)~\cite{10156812Wang} and diffusion models like Stable Diffusion~\cite{Rombach_2022_CVPR} have made it possible to create highly realistic images with minimal input from users. The accessibility of these tools has expanded, with open-source implementations and user-friendly interfaces making them available to a broader audience. While these developments have numerous beneficial applications in fields such as entertainment, art, and design, they also pose significant risks. The ease with which realistic images can be generated raises concerns about their potential misuse in activities like blackmail, manipulation, and the spread of misinformation~\cite{Vaccari2020DeepfakesAD}. 

Detecting AI-generated images has thus become an essential area of research. Convolutional Neural Networks (CNNs) have shown promise in image classification tasks, including the detection of manipulated or synthesized images~\cite{bird2023cifakeimageclassificationexplainable}. However, as AI-generated images become more realistic, existing detection methods require enhancement to maintain their effectiveness.

This paper focuses on evaluating the effectiveness of using CNNs to detect AI-generated images, particularly those produced by Stable Diffusion-based generators. We explore vulnerabilities in current detection approaches, such as susceptibility to adversarial attacks and overfitting to specific data distributions. Additionally, we investigate the use of DenseNet architectures~\cite{huang2018denselyconnectedconvolutionalnetworks} to improve the accuracy and robustness of detecting AI-generated images. DenseNets, known for their efficient feature propagation and reduced parameter count, may offer advantages over traditional CNNs in this context.

The datasets used in this study are variations of the CIFAKE dataset~\cite{huggingfaceDragonintelligenceCIFAKEimagedatasetDatasets}, a widely referenced resource for training and evaluating AI-image detectors. The original CIFAKE dataset comprises 120,000 images: 60,000 real images sourced from the CIFAR-10 dataset and 60,000 synthetic images generated using Stable Diffusion 1.4. The synthetic images replicate the categories in CIFAR-10 (e.g., airplanes, cats, and trucks) using prompts such as "A photograph of [object]," supplemented with context-specific modifiers to enhance realism. All images were resized to 32x32 pixels for computational efficiency. To assess the generalizability and limitations of the CNN-based models, two extended datasets—CIFAKE-SD2.1 and CIFAKE-SD3.0—were created using Stable Diffusion 2.1 and 3.0, respectively. These datasets preserve the structure and composition of the original CIFAKE dataset but feature AI-generated images from updated versions of Stable Diffusion, providing a robust testbed for evaluating the impact of model updates on detection accuracy. 

In addition to examining the impact of different Stable Diffusion versions on model detection accuracy, this study investigates various other factors that could influence detection performance, including alterations and modifications to image generation or the generated images themselves. Specifically, factors such as Gaussian blurring, variations in prompt text, and adjustments to the image generation model using Low-Rank Adaptation (LoRA)~\cite{hu2021loralowrankadaptationlarge} are analyzed. Corresponding datasets, including CIFAKE-SD2.1-Blurred, CIFAKE-SD2.1-GPT4o, and CIFAKE-SD2.1-LoRA, were systematically generated to facilitate these evaluations.


\section{Related Work}
The rapid advancement of generative models has led to the proliferation of highly realistic AI-generated images, raising concerns about authenticity and the potential for misuse. Detecting these synthetic images has become a critical area of research, with various methodologies proposed to address the challenge. Convolutional Neural Networks (CNNs) have been extensively employed for image classification and forgery detection tasks. For instance, \cite{huang2018denselyconnectedconvolutionalnetworks} introduced the DenseNet architecture, which enhances feature propagation and reduces redundancy by connecting each layer to every other layer in a feed-forward manner. This architecture has proven effective in image recognition tasks due to its efficient use of parameters and improved gradient flow. Building on the strengths of CNNs, \cite{bird2023cifakeimageclassificationexplainable} proposed a CNN-based approach specifically for detecting AI-generated images, demonstrating notable accuracy in distinguishing synthetic content. Generative Adversarial Networks (GANs) have been at the forefront of generating realistic images. The work by \cite{10156812Wang} delves into the mechanisms of GANs, highlighting how the generator and discriminator networks compete to produce lifelike images. While GANs excel in image synthesis, they also present challenges in detection due to the high quality of generated images. To address the evolving sophistication of generative models, \cite{10431766Alhabeeb} provided a comprehensive review of text-to-image synthesis techniques, including GANs and diffusion models. The paper compares various models, discussing their advantages and limitations, and underscores the need for robust detection methods as generative models continue to improve. Image forgery detection has also been approached through the analysis of compression artifacts. \cite{10205377Patel} introduced a method combining Error Level Analysis (ELA) and CNNs to identify inconsistencies in image compression levels, effectively detecting manipulated images. This technique leverages the fact that edited regions often exhibit different compression characteristics compared to the rest of the image. In terms of protecting the integrity and ownership of AI-generated images, watermarking techniques have been explored. \cite{10612834Yuan} proposed embedding watermarks into Stable Diffusion Models (SDMs) to assert ownership and safeguard intellectual property. Their method involves fine-tuning the SDM to generate specific watermarks in response to predefined prompts, thereby proving model ownership without compromising performance. Robustness of detection methods under image alterations is another critical aspect. \cite{10508937Park} conducted a performance comparison of AI-generated image detection methods, evaluating their resilience to image manipulations such as JPEG compression and Gaussian blurring. They utilized tools like Grad-CAM and t-SNE for visualization, providing insights into the methods' effectiveness under challenging conditions.

\subsection{CIFAKE Dataset and Classifier}

Bird and Lotfi~\cite{bird2023cifakeimageclassificationexplainable} introduced the CIFAKE dataset and proposed a Convolutional Neural Network (CNN) to classify images as either real or AI-generated. The classifier processes 32x32 pixel RGB images and outputs a binary decision, with values above 0.5 classified as real. The optimal network architecture comprises two convolutional layers with 32 filters each and two fully connected layers, achieving an accuracy of 92.93\% with a binary cross-entropy loss of 0.18. Despite its success, the study did not provide details on key training parameters, such as optimizers and learning rates, leaving room for further exploration.

The CIFAKE dataset includes 120,000 images, evenly split between real and AI-generated categories. The real images are sourced from CIFAR-10, spanning 10 categories such as airplanes and cats. The AI-generated images were created using Stable Diffusion 1.4 with prompts like “A photograph of [object],” along with category-specific modifiers, and resized to 32x32 pixels for consistency. This dataset has been instrumental in evaluating the performance of detection models under controlled conditions.

\begin{figure}[t]
\vspace{-2 ex}
      \centering
      \includegraphics[width=\columnwidth]{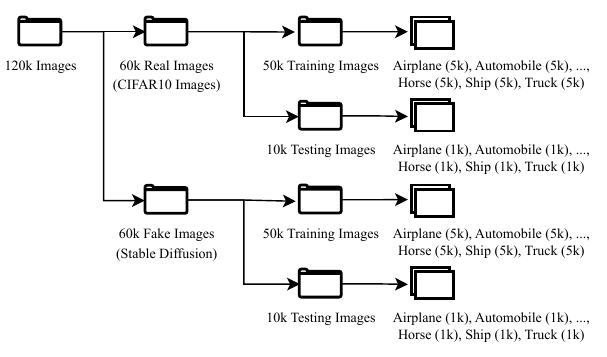}
      \caption{CIFAKE Dataset structure.}
      \label{fig:cifakestruct}
      \vspace{-2 ex}
\end{figure}

This paper builds upon CIFAKE by replicating its methodology and further evaluating the effectiveness and vulnerabilities of the proposed CNN-based classifier. Specifically, this research examines its robustness against variations in image generation, such as newer versions of Stable Diffusion, Gaussian blurring, and modifications using Low-Rank Adaptation (LoRA). The CIFAKE study provides the foundational framework and motivation for this work, enabling a deeper investigation into the resilience and limitations of AI-generated image detectors.

\begin{figure*}[t]
    \centering
    \includegraphics[width=\textwidth]{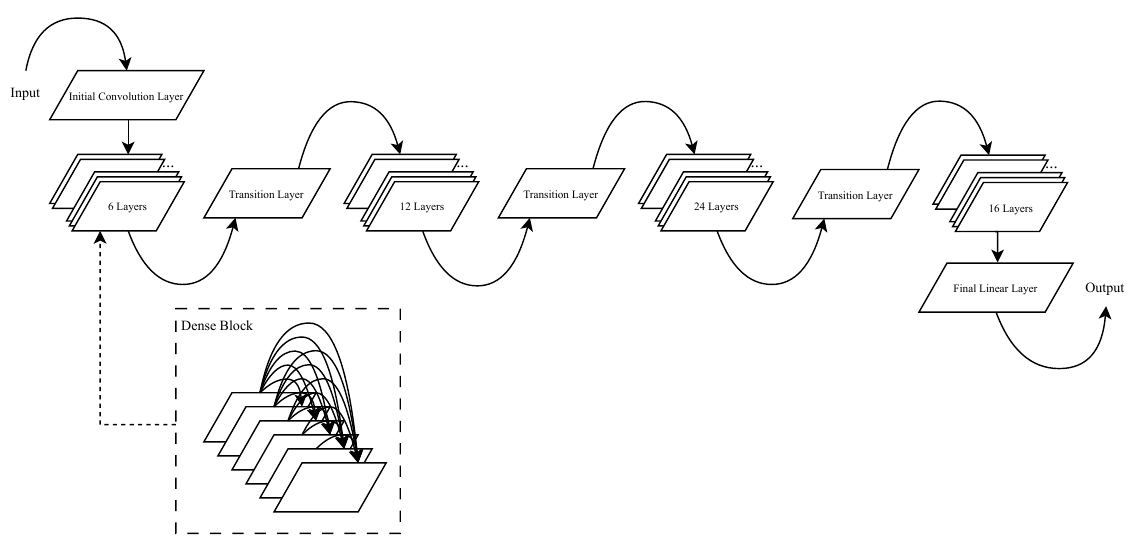} 
    \caption{Structure of the DenseNet model used in our experiments. The model includes an initial convolutional layer, followed by multiple dense blocks and transition layers, concluding with a final linear layer for binary classification.}
    \label{fig:densenet_structure}
\end{figure*}

\section{Methods}

To evaluate and enhance the effectiveness of AI-generated image classifiers, this study first focused on generating a diverse set of datasets to comprehensively test the robustness of classifiers under various conditions. In real-world application scenarios, AI-generated image classifier services typically have no control over the methods used to generate the provided images. To simulate such unconstrained scenarios, we generated several datasets by introducing diverse alternations to the original Stable Diffusion image generation methods.

For datasets representing outputs from advanced AI models, newer versions of Stable Diffusion were utilized, reflecting the evolution of AI generation capabilities over time. To mimic real-world image degradation, Gaussian blurring was applied to artificially introduce imperfections such as those caused by camera focus or resolution issues. To simulate scenarios where bad actors use fine-tuned Stable Diffusion models, datasets were generated with Stable Diffusion fine-tuned via Low-Rank Adaptation (LoRA)~\cite{hu2021loralowrankadaptationlarge}, allowing for the creation of highly photorealistic images with reduced detectable artifacts. Additionally, recognizing that the original CIFAKE dataset was limited by a fixed set of prompts, we introduced datasets generated using a broader and more diverse set of prompts created by large language models. This allowed us to evaluate vulnerabilities related to overfitting on fixed prompt templates and expose potential weaknesses when classifiers encounter unseen prompt variations.

These varied dataset generation approaches ensured a comprehensive evaluation of the classifiers' robustness and their ability to adapt to challenging scenarios and diverse inputs.

Building on the foundation provided by the Nottingham-Trent CNN-based classifier, we also explored ways to improve its performance and robustness. A DenseNet121 architecture~\cite{huang2018denselyconnectedconvolutionalnetworks} was adapted for this task, leveraging its densely connected layers to enhance gradient flow, improve feature reuse, and reduce the number of trainable parameters compared to traditional CNNs of similar depth. DenseNet's architecture is particularly suited for image classification tasks where capturing fine-grained patterns and preserving information across layers are critical.

To accommodate the computational constraints of this study and the dataset characteristics, the DenseNet model was modified to process 32x32 pixel images, similar to the input size used in the CIFAKE dataset, and was tailored for binary classification tasks. The modified architecture retained DenseNet’s core strengths while adapting to the dataset's size and format, shown as Figure~\ref{fig:densenet_structure}.

The general approach involved training and testing both CNN and DenseNet models on the original CIFAKE dataset and the newly generated alternated datasets. This two-pronged strategy—first creating challenging datasets to test the classifiers' resilience and then introducing an advanced neural network architecture—enabled a thorough evaluation of the original classifier and allowed for the proposal of meaningful improvements. This methodology not only addresses vulnerabilities in the Nottingham-Trent classifier but also establishes a benchmark for detecting increasingly sophisticated AI-generated images in dynamic, real-world scenarios.

To replicate the CNN-based detection approach proposed by Bird and Ahmad Lotfi~\cite{bird2023cifakeimageclassificationexplainable}, the neural network design was implemented with reasonable assumptions to fill gaps in the original description. The architecture included two convolutional layers, each with 32 filters, followed by two linear layers. Missing details, such as the kernel size, padding, and pooling configuration, were addressed by selecting commonly used values: a square kernel size of 3 with a padding of 1 for the convolutional layers and a kernel size of 2 with a stride of 2 for pooling layers. The output layer applied a Sigmoid activation function for binary classification, while ReLU was used for the intermediate layers. The implementation utilized the Adam optimizer~\cite{kingma2017adammethodstochasticoptimization} to handle sparse gradients and improve convergence. Hyperparameters included a learning rate of $10^{-3}$, a batch size of 1,000, and training durations of 5, 10, and 15 epochs to explore the impact of training time on performance.

\section{Experiments}
\renewcommand{\arraystretch}{1.06}
\setlength{\tabcolsep}{1.0em}
\begin{table*}[ht]
\centering
\caption{Summary of Datasets Used in Experiments}
\label{table:datasets}
\begin{tabular}{l p{12cm}}
\hline
\textbf{Dataset Name} & \textbf{Description} \\
\hline
\textbf{CIFAKE} & Original dataset used in the Nottingham Trent University paper; consists of 60,000 real images from CIFAR-10 and 60,000 AI-generated images created using Stable Diffusion 1.4 with the prompt "A photograph of a/an ...". \\[2ex]

\textbf{CIFAKE-SD2.1} & Similar to CIFAKE but AI-generated images were created using Stable Diffusion 2.1; images generated at 512\(\times\)512 pixels. \\[2ex]

\textbf{CIFAKE-SD3.0} & Similar to CIFAKE but AI-generated images were created using Stable Diffusion 3.0; images generated at 512\(\times\)512 pixels. \\[2ex]

\textbf{CIFAKE-SD2.1-Blurred} & CIFAKE-SD2.1 dataset where images were Gaussian blurred with a radius of 5 pixels and standard deviation \(\sigma = 1.1\). \\[2ex]

\textbf{CIFAKE-SD2.1-768} & AI-generated images were created using Stable Diffusion 2.1 at 768\(\times\)768 pixels (default for SD 2.1). \\[2ex]

\textbf{CIFAKE-SD3.0-1024} & AI-generated images were created using Stable Diffusion 3.0 at 1024\(\times\)1024 pixels (default for SD 3.0). \\[2ex]

\textbf{CIFAKE-SD2.1-P2} & Similar to CIFAKE-SD2.1 but AI-generated images were created using the prompt "A photo of ..., real". \\[2ex]

\textbf{CIFAKE-SD2.1-P3} & Similar to CIFAKE-SD2.1 but AI-generated images were created using the prompt "Realistic photo of ...". \\[2ex]

\textbf{CIFAKE-SD2.1-GPT4o} & AI-generated images were created using highly specific prompts generated by OpenAI's GPT-4o model. \\[2ex]

\textbf{CIFAKE-SD2.1-Negative} & Similar to CIFAKE-SD2.1 but added negative prompts to avoid certain characteristics like "blurry, distorted, low quality, etc.". \\[2ex]

\textbf{CIFAKE-SD2.1-LoRA} & AI-generated images were created using Stable Diffusion 2.1 tuned with Low-Rank Adaptation (LoRA); LoRA was trained using the MIT-Adobe FiveK dataset~\cite{fivek}. \\
\hline
\end{tabular}
\end{table*}

\begin{figure}[t]
    \centering

    \begin{subfigure}[b]{\linewidth}
        \centering
        \includegraphics[width=0.9\linewidth]{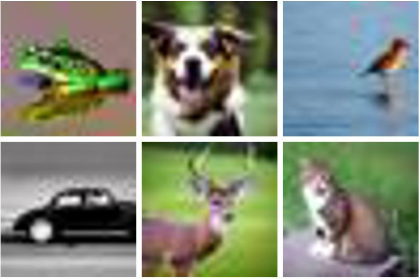} 
        \caption{6 random stable diffusion-generated images from CIFAKE (32$\times32$).}
        \label{fig:cifake_generated}
    \end{subfigure}

    \vspace{0.5cm} 

    \begin{subfigure}[b]{\linewidth}
        \centering
        \includegraphics[width=0.9\linewidth]{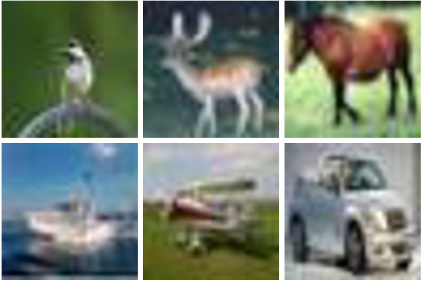} 
        \caption{6 random real images from CIFAKE (32$\times32$).}
        \label{fig:cifake_real}
    \end{subfigure}

    \caption{Example images from the CIFAKE~\cite{huggingfaceDragonintelligenceCIFAKEimagedatasetDatasets} dataset. \label{fig:cifakesamples}}
\end{figure}

\label{sec:results}
\label{sec:experiment-results}
To thoroughly evaluate the robustness and effectiveness of the classifiers, we conducted a series of experiments across multiple datasets, including the original CIFAKE dataset and its alternated versions generated with diverse approaches. {Code and is available at \url{https://github.com/JustinJiangNext/AI-Image-Detection-Benchmarking}. Datasets are available at \url{https://www.kaggle.com/justinjiangnext/datasets}}.

\subsection{Datasets}

\begin{table}[t]
\centering
\caption{Performance of the CNN model at different training epochs, where BCE refers to Binary Cross-Entropy Loss.}
\label{table:cifakereplication}
\begin{tabular}{c c p{3cm}}
\hline
\textbf{Epochs} & \textbf{Accuracy (\%)} & \textbf{BCE Loss} \\ \hline
5               & 90.47                  & 0.2352       \\
10              & 90.83                  & 0.2033       \\
15              & 93.67                  & 0.1706       \\ \hline
\end{tabular}
\end{table}

To extend the evaluation, two additional datasets were created using newer versions of Stable Diffusion: CIFAKE-SD2.1 and CIFAKE-SD3.0. These datasets maintained the same structure as CIFAKE but used Stable Diffusion 2.1~\cite{huggingfaceStabilityaistablediffusion21Hugging} and 3.0~\cite{huggingfaceStabilityaistablediffusion3mediumHugging} to generate the synthetic images. Additional datasets were generated to test specific vulnerabilities, including: CIFAKE-SD2.1-Blurred: Gaussian-blurred images to obscure detection-relevant patterns;
CIFAKE-SD2.1-P2 and CIFAKE-SD2.1-P3: Images generated with slightly altered prompts to test sensitivity to prompt variations;
CIFAKE-SD2.1-GPT4o: Images generated using highly specific prompts created by GPT-4;
CIFAKE-SD2.1-Negative: Images generated with negative prompts to suppress visual artifacts; and
CIFAKE-SD2.1-LoRA: Images generated using Stable Diffusion fine-tuned with Low-Rank Adaptation (LoRA) for enhanced photorealism.
A completed list of datasets can be found in Table~\ref{table:datasets}.

\subsection{Results}
The results of this study are presented to evaluate the performance, robustness, and limitations of AI-generated image classifiers under various experimental conditions, including modifications to datasets and model architectures.

\vspace{-2ex}
\paragraph{Replicating CIFAKE Method.}
In order to establish a reliable baseline for subsequent experiments, this paper evaluated the CNN model’s accuracy and binary cross-entropy loss across different training durations (5, 10, and 15 epochs) while keeping all other variables constant. As shown in Table~\ref{table:cifakereplication}, the model trained for 15 epochs achieved an accuracy of 93.67\% and a binary cross-entropy loss of 0.1706, closely matching the 92.93\% accuracy and 0.18 loss reported in the original study. This alignment validates the robustness of the original approach. Consequently, all subsequent experiments in this paper were conducted using models trained for 15 epochs. Notably, despite the reduced resolution of 32$\times$32 pixels used for classification, the classifier successfully distinguished between real and Stable Diffusion-generated images, which remain indistinguishable to human vision.

\vspace{-2ex}
\paragraph{Evaluation Across Stable Diffusion Versions.}
The original CIFAKE dataset’s AI-generated component was created using Stable Diffusion 1.4 (Diffusers, trained by CompVis~\cite{Rombach_2022_CVPR}). Since the publication of the study, newer versions of Stable Diffusion, specifically 2.1 and 3.0, have been released, offering enhanced image generation capabilities. To evaluate whether the CNN model trained on Stable Diffusion 1.4 would perform less accurately on images generated by these newer versions, we generated two additional datasets: CIFAKE-SD2.1 and CIFAKE-SD3.0.

Both CIFAKE-SD2.1 and CIFAKE-SD3.0 followed the structure of the original CIFAKE dataset, comprising 60,000 real images from the CIFAR10 dataset~\cite{Krizhevsky09learningmultiple} and 60,000 AI-generated images. The primary difference lay in the Stable Diffusion version used for generating synthetic images, which incorporated models developed by Stability AI through the Diffusers library. While Stable Diffusion 1.4 produces images of size 512$\times$512 pixels by default, the newer versions—2.1 and 3.0—generate images with higher resolutions of 768$\times$768 and 1024$\times$1024 pixels, respectively. For consistency, all synthetic images in CIFAKE-SD2.1 and CIFAKE-SD3.0 were resized to 512$\times$512 pixels before downscaling to 32x32 for training and evaluation.

To test the impact of version differences, we trained three CNN models on CIFAKE, CIFAKE-SD2.1, and CIFAKE-SD3.0, and then evaluated their performance on all three datasets. The results are summarized in Table~\ref{table:combined_performance}. The accuracy table shows how well each model performed across both real and AI-generated images, while the second table focuses on how effectively the models identified AI-generated images.

\begin{table*}[t]
\centering
\caption{Performance of CNN models across different datasets. (a) Overall accuracy includes both real and AI-generated images. (b) Fake image accuracy represents the model's ability to identify AI-generated images specifically.}
\label{table:combined_performance}
\begin{tabular}{c c c c}
\multicolumn{4}{c}{\textbf{(a) Overall Accuracy (\%)}} \\ \hline
\textbf{Trained Model} & \textbf{CIFAKE (\%)} & \textbf{CIFAKE-SD2.1 (\%)} & \textbf{CIFAKE-SD3.0 (\%)} \\ \hline
CIFAKE                 & 93.67                & 92.30                      & 91.21                      \\ 
CIFAKE-SD2.1           & 81.89                & 95.23                      & 84.30                      \\ 
CIFAKE-SD3.0           & 70.28                & 81.64                      & 96.84                      \\ \hline
\end{tabular}

\vspace{1em} 

\begin{tabular}{c c c c}
\multicolumn{4}{c}{\textbf{(b) Fake Image Accuracy (\%)}} \\ \hline
\textbf{Trained Model} & \textbf{CIFAKE (\%)} & \textbf{CIFAKE-SD2.1 (\%)} & \textbf{CIFAKE-SD3.0 (\%)} \\ \hline
CIFAKE                 & 93.23                & 90.48                      & 88.30                      \\ 
CIFAKE-SD2.1           & 71.42                & 98.10                      & 76.24                      \\ 
CIFAKE-SD3.0           & 44.63                & 67.34                      & 97.75                      \\ \hline
\end{tabular}
\end{table*}

\begin{table*}[t]
\centering
\caption{Performance of CNN models on blurred datasets. (a) Overall accuracy of the CNN model trained on CIFAKE-SD2.1 and tested on CIFAKE-SD2.1-Blurred. (b) Fake image accuracy of the CNN model trained on CIFAKE-SD2.1 and tested on CIFAKE-SD2.1-Blurred.}
\label{table:combined_performance_blur}
\begin{tabular}{c c c}
\multicolumn{3}{c}{\textbf{(a) Overall Accuracy (\%)}} \\ \hline
\textbf{Trained Model} & \textbf{CIFAKE-SD2.1 (\%)} & \textbf{CIFAKE-SD2.1-Blurred (\%)} \\ \hline
CIFAKE-SD2.1           & 95.23                      & 71.13                               \\ \hline
\end{tabular}

\vspace{1em} 

\begin{tabular}{c c c}
\multicolumn{3}{c}{\textbf{(b) Fake Image Accuracy (\%)}} \\ \hline
\textbf{Trained Model} & \textbf{CIFAKE-SD2.1 (\%)} & \textbf{CIFAKE-SD2.1-Blurred (\%)} \\ \hline
CIFAKE-SD2.1           & 98.10                      & 49.90                               \\ \hline
\end{tabular}
\end{table*}

\begin{table*}[t]
\centering
\caption{Accuracy of CNN models trained on 512$\times$512 images and tested on datasets generated at larger default resolutions (768$\times$768 for CIFAKE-SD2.1 and 1024$\times$1024 for CIFAKE-SD3.0).}
\label{table:image_size_sensitivity}
\begin{tabular}{c c c}
\hline
\textbf{Trained Model} & \textbf{Original Size Accuracy (\%)} & \textbf{Larger Size Accuracy (\%)} \\ \hline
CIFAKE-SD2.1           & 95.23                               & 93.90                              \\ 
CIFAKE-SD3.0           & 96.84                               & 88.76                              \\ \hline
\end{tabular}
\end{table*}

\begin{table*}[t]
\centering
\caption{Accuracy of CNN models trained and tested on datasets with similar prompts.}
\label{table:prompt_similarity}
\begin{tabular}{c c c c}
\hline
\textbf{Trained Model} & \textbf{CIFAKE-SD2.1 (\%)} & \textbf{CIFAKE-SD2.1-P2 (\%)} & \textbf{CIFAKE-SD2.1-P3 (\%)} \\ \hline
CIFAKE-SD2.1           & 95.23                      & 95.36                          & 95.26                          \\
CIFAKE-SD2.1-P2        & 96.02                      & 96.47                          & 96.03                          \\
CIFAKE-SD2.1-P3        & 95.65                      & 96.04                          & 96.18                          \\ \hline
\end{tabular}
\end{table*}

\begin{table*}[t]
\centering
\caption{Accuracy of the CNN model on CIFAKE-SD2.1, CIFAKE-SD2.1-GPT4o, and CIFAKE-SD2.1-Negative datasets with specific and negative prompts.}
\label{table:prompt_accuracy}
\begin{tabular}{c c c c}
\hline
\textbf{Trained Model} & \textbf{CIFAKE-SD2.1 (\%)} & \textbf{CIFAKE-SD2.1-GPT4o (\%)} & \textbf{CIFAKE-SD2.1-Negative (\%)} \\ \hline
CIFAKE-SD2.1           & 95.23                      & 93.78                             & 94.06                                \\ \hline
\end{tabular}
\end{table*}

\begin{table*}[t]
\centering
\caption{Performance of CNN models on LoRA-alternated datasets. (a) Overall accuracy of the CNN model trained on CIFAKE-SD2.1 and tested on CIFAKE-SD2.1-LoRA. (b) Fake image accuracy of the CNN model trained on CIFAKE-SD2.1 and tested on CIFAKE-SD2.1-LoRA.}
\label{table:lora_performance}
\begin{tabular}{c c c}
\multicolumn{3}{c}{\textbf{(a) Overall accuracy (\%)}} \\ \hline
\textbf{Trained Model} & \textbf{CIFAKE-SD2.1 (\%)} & \textbf{CIFAKE-SD2.1-LoRA (\%)} \\ \hline
CIFAKE-SD2.1           & 95.23                      & 85.27                           \\ \hline
\end{tabular}

\vspace{1em} 

\begin{tabular}{c c c}
\multicolumn{3}{c}{\textbf{(b) Fake Image Accuracy (\%)}} \\ \hline
\textbf{Trained Model} & \textbf{CIFAKE-SD2.1 (\%)} & \textbf{CIFAKE-SD2.1-LoRA (\%)} \\ \hline
CIFAKE-SD2.1           & 98.10                      & 78.18                           \\ \hline
\end{tabular}
\end{table*}

\begin{table*}[t]
\centering
\caption{Comparison of DenseNet and CNN performance (trained on CIFAKE-SD2.1) on real and AI-generated images across CIFAKE, CIFAKE-SD2.1, and CIFAKE-SD3.0 datasets.}
\label{table:densenet_cnn_accuracy_part1}
\begin{tabular}{c c c c}
\hline
\textbf{Model}         & \textbf{CIFAKE (\%)} & \textbf{CIFAKE-SD2.1 (\%)} & \textbf{CIFAKE-SD3.0 (\%)} \\ \hline
DenseNet               & 97.23                & 98.57   & 98.78 \\
CNN                    & 93.67                & 95.23   & 96.84                   \\ \hline
\end{tabular}
\end{table*}

\begin{table*}[t]
\centering
\caption{Comparison of DenseNet and CNN performance (trained on CIFAKE-SD2.1) on Gaussian blurred datasets.}
\label{table:densenet_cnn_accuracy_part2}
\begin{tabular}{c c c}
\hline
\textbf{Model}         & \textbf{Blurred Accuracy (\%)} & \textbf{Blurred Fake Accuracy (\%)} \\ \hline
DenseNet               & 86.88                          & 75.18                               \\
CNN                    & 71.13                          & 49.90                               \\ \hline
\end{tabular}
\end{table*}

\vspace{-2ex}
\paragraph{Evaluation on Gaussian Blur.}
To explore if the CNN model relies on specific version-dependent patterns of Stable Diffusion-generated images, this paper tested its performance on a modified dataset where these patterns were obscured through Gaussian blurring. This experiment aimed to simulate real-world imperfections such as focus issues or resolution degradation, which are commonly introduced by adversarial or accidental manipulations. The resulting dataset, CIFAKE-SD2.1-Blurred, retains the structure of CIFAKE-SD2.1 but applies Gaussian blur to all images.

Gaussian blurring was chosen for its ability to degrade image quality smoothly and realistically, mimicking conditions that make it harder for both humans and models to distinguish between real and AI-generated images. The blur was applied with a radius of 5 pixels and a standard deviation ($\sigma$) of 1.1, calculated using OpenCV's~\cite{itseez2015opencv} formula:

\begin{equation}
\sigma=0.3\left(\frac{kernel\_size - 1}{2}-1\right)
\end{equation}

This ensures the blurring effect remains natural and does not introduce artifacts or visual distortions.

The CIFAKE-SD2.1-trained CNN model was tested on both the CIFAKE-SD2.1 and CIFAKE-SD2.1-Blurred datasets to assess its robustness. The accuracy results are presented in Table~\ref{table:combined_performance_blur}.

\vspace{-2ex}
\paragraph{Evaluation on Image Size Sensitivity.}
The default image sizes generated by Stable Diffusion versions differ, with Stable Diffusion 2.1 producing 768$\times$768 pixels and Stable Diffusion 3.0 generating 1024$\times$1024 pixels. In previous experiments, images were standardized to 512$\times$512 pixels to align with the default size of Stable Diffusion 1.4 and to isolate the Stable Diffusion version as the primary variable. However, this standardization might obscure potential model sensitivities to variations in image size.

To evaluate the impact of image size on classification accuracy, additional datasets—CIFAKE-SD2.1-768 and CIFAKE-SD3.0-1024—were generated. CIFAKE-SD2.1-768 contains images generated using Stable Diffusion 2.1 at its default resolution of 768$\times$768 pixels, while CIFAKE-SD3.0-1024 includes images generated by Stable Diffusion 3.0 at 1024x1024 pixels. The CNN model trained on CIFAKE-SD2.1 was tested on CIFAKE-SD2.1-768, and the model trained on CIFAKE-SD3.0 was tested on CIFAKE-SD3.0-1024.

The results are presented in Table~\ref{table:image_size_sensitivity}, showing the overall accuracy of the models when tested on datasets with varying image sizes.

\vspace{-2ex}
\paragraph{Evaluation on Prompt Variability.} 
The CIFAKE dataset relies on a fixed and limited set of prompts to generate its AI-generated image component, specifically in the format "A photograph of a/an ...". This approach may fail to represent the diverse and flexible ways prompts can be structured in real-world applications. Bad actors are likely to manipulate prompts to produce images that are more challenging to detect. To address this limitation, a series of experiments was conducted using datasets generated with varied and more specific prompts to evaluate the robustness of the CNN model.

To test the effects of slight variations in prompts, two additional datasets were created: CIFAKE-SD2.1-P2, using the prompt "A photo of ..., real," and CIFAKE-SD2.1-P3, using the prompt "Realistic photo of ...". These datasets maintained the same structure and resolution (512$\times$512 pixels) as CIFAKE-SD2.1. The accuracy results are summarized in Table~\ref{table:prompt_similarity}.

To simulate real-world scenarios where prompts may be more detailed, the CIFAKE-SD2.1-GPT4o dataset was created. This dataset used 125 unique prompts per category, generated using OpenAI's GPT-4o, resulting in a total of 60,000 AI-generated images. Prompts included detailed scenarios like "A plane flying low over a beach with sunbathers watching" or "A commercial airplane parked at an airport gate at night." Table~\ref{table:prompt_accuracy} shows the accuracy results.

Another potential adversarial tactic involves using negative prompts to refine AI-generated images, avoiding characteristics that might make them detectable as fake. To test this, the CIFAKE-SD2.1-Negative dataset was created by adding negative prompts to avoid traits such as "blurry, distorted, low quality, surreal, or cartoonish." Table~\ref{table:prompt_accuracy} presents the results.

\vspace{-2ex}
\paragraph{Evaluation on  Low-Rank Adaptation (LoRA) Alternation.} 
Stable Diffusion can produce a wide range of images, from illustrations and art to highly photorealistic imagery. One method to tune Stable Diffusion for specific styles or themes is Low-Rank Adaptation (LoRA), a fine-tuning technique that enables targeted modifications without requiring extensive computational resources. LoRA modifies the pre-trained model's architecture by introducing low-rank matrices, which capture task-specific features while preserving the broader generative capabilities of the original model~\cite{hu2021loralowrankadaptationlarge}. This technique is particularly beneficial for adapting Stable Diffusion to generate highly photorealistic images, potentially avoiding detectable "fingerprinting" patterns.

\begin{figure}[t]
\centering
\includegraphics[width=0.45\textwidth]{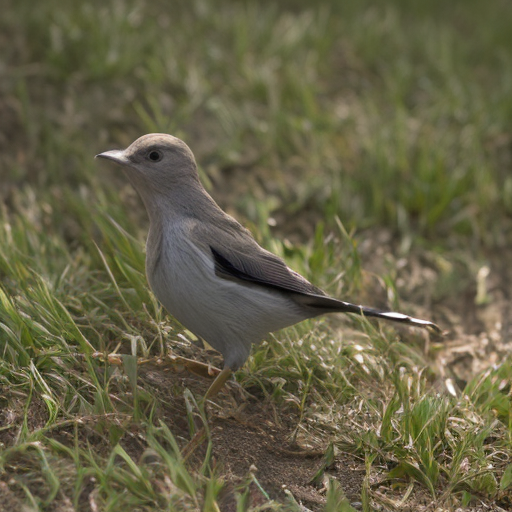}
\caption{Example image from the CIFAKE-SD2.1-LoRA dataset with its photorealism trigger word.}
\label{fig:lora_example}
\end{figure}

To evaluate the impact of LoRA tuning on Stable Diffusion and the CNN model's performance, we used the MIT-Adobe FiveK dataset to train a photorealism-oriented LoRA. The dataset comprises 5,000 images labeled with detailed textual descriptions excluding general quality characteristics, such as lighting conditions. Labels were generated using the Llama 3.2 Vision Instruct model, combined with a unique trigger word \textit{R3E4AL}, representing photorealism. For example, an image of a wooden structure in a forest was labeled as "R3E4AL, a decaying wooden structure in a green forest with grass" (Figure~\ref{fig:lora_example}).

After training the LoRA, it was integrated into Stable Diffusion 2.1 to create the CIFAKE-SD2.1-LoRA dataset. This dataset was then used to evaluate the CNN model trained on CIFAKE-SD2.1. The performance of the CNN model trained on CIFAKE-SD2.1 and evaluated on CIFAKE-SD2.1-LoRA is summarized in Table~\ref{table:lora_performance}. The overall accuracy dropped significantly from 95.23\% to 85.27\%. When evaluating the model's ability to detect AI-generated images specifically, the accuracy dropped further to 78.18\%, as shown in Table~\ref{table:lora_performance}. This indicates that the LoRA-tuned Stable Diffusion images appear sufficiently realistic to confuse the classifier.

\vspace{-2ex}
\paragraph{Using DenseNets to Detect Stable Diffusion-Generated Images.} 
DenseNet (Dense Convolutional Network) offers an advanced deep learning architecture that enhances feature reuse and gradient flow, making it especially effective for image processing tasks. Unlike traditional convolutional neural networks (CNNs), DenseNet employs dense connectivity, connecting each layer to all subsequent layers. This approach ensures that early layers’ features, such as edges and textures, are readily accessible to later layers, enabling better decision-making and reducing redundancy.

DenseNet consists of dense blocks and transition layers. Each dense block contains several layers with outputs concatenated to all preceding layers, allowing the model to retain detailed representations. Transition layers compress accumulated features using convolution and pooling, balancing computational efficiency with detailed feature retention. For this study, the DenseNet121 architecture was selected and modified to process 32$\times$32 pixel images and output a single binary classification value. Adjustments included reducing the kernel size, stride, and padding in the first convolution layer to match the input size and modifying the final layer to output a single classification value.

\section{Discussion}
The findings highlight both the strengths and vulnerabilities of current AI-generated image classification methods, offering insights into their practical applications and areas for improvement in handling diverse and evolving generative techniques.

\vspace{-2ex}
\paragraph{Stable Diffusion Version Overfitting.}
From these results, three key observations emerge. First, models trained on a dataset generated with a specific version of Stable Diffusion performed poorly when tested on datasets generated with different versions, highlighting a lack of generalization. Second, models struggled significantly to identify images generated by older versions of Stable Diffusion, with substantial accuracy drops when tested on earlier datasets. Finally, models trained on newer versions (e.g., CIFAKE-SD3.0) performed better in identifying images generated by their respective versions compared to those trained on older versions.

Interestingly, while images from newer versions of Stable Diffusion appear more realistic to humans, the CNN model demonstrated better accuracy in detecting these images compared to older versions. This suggests that newer Stable Diffusion models introduce more distinct patterns or "fingerprints," which the CNN classifier can more effectively learn. However, the sensitivity of CNN models to version-specific patterns underscores a significant limitation: their reliance on consistent generative models and their vulnerability to evolving AI generation techniques. This observation emphasizes the need for classifiers that generalize well across varying generative methods.

\vspace{-2ex}
\paragraph{Gaussian  Blur.}
From Table~\ref{table:combined_performance_blur}, the CNN model's overall accuracy drops significantly from 95.23\% to 71.13\% when tested on blurred images. More notably, Table~\ref{table:combined_performance_blur}(b) reveals that the fake image accuracy plummets to 49.90\%, which is effectively random guessing for a binary classification problem. These results confirm that Gaussian blurring disrupts the patterns or "fingerprints" the CNN model relies upon for detection.

This experiment highlights the vulnerability of CNN-based classifiers to simple image modifications like Gaussian blurring. Such manipulations can significantly impair the classifier's ability to identify AI-generated images, underlining the need for more robust detection models capable of adapting to diverse real-world scenarios.

\vspace{-2ex}
\paragraph{Image Size Sensitivity.}
From Table~\ref{table:image_size_sensitivity}, it is evident that image size has a moderate effect on model accuracy. The CNN model trained on CIFAKE-SD2.1 showed a relatively minor accuracy drop of 1.33 percentage points when tested on CIFAKE-SD2.1-768. In contrast, the CNN model trained on CIFAKE-SD3.0 exhibited a more significant decrease of 8.08 percentage points when tested on CIFAKE-SD3.0-1024.

This discrepancy suggests that larger differences in image resolution between training and testing datasets may disproportionately affect model performance. The model trained on CIFAKE-SD3.0 may be more impacted due to the greater disparity between the training (512$\times$512) and testing (1024$\times$1024) image sizes. These findings highlight the importance of considering image resolution consistency in training and evaluation pipelines for AI-generated image classifiers.

\vspace{-2ex}
\paragraph{Prompt Variability.}
The results in Table~\ref{table:prompt_similarity} indicate that minor modifications to the prompt have little to no effect on the model's accuracy. This suggests that the CNN model generalizes well to prompt variations within a similar structure. Although the accuracy slightly decreased when tested on more specific prompts, The results in Table~\ref{table:prompt_accuracy} the CNN model still demonstrated robustness, achieving an accuracy of 93.78\%. This indicates that while specific prompts introduce some variability, they do not significantly impair the model's ability to detect AI-generated images. The results in Table~\ref{table:prompt_accuracy} also reveal only a slight drop in accuracy, from 95.23\% to 94.06\%, indicating that negative prompts have minimal impact on the CNN model's ability to detect AI-generated images. This further supports the hypothesis that the model relies on intrinsic "fingerprints" of Stable Diffusion rather than visual defects detectable by humans.

The experiments with varied prompts demonstrate the robustness of the CNN model against different prompt structures and strategies, including both slight modifications and highly specific or adversarial prompt designs. This suggests that the model primarily leverages stable and consistent patterns inherent to AI-generated images rather than relying on superficial prompt-related cues.

\vspace{-2ex}
\paragraph{Low-Rank Adaptation (LoRA) and Stable Diffusion.}
The results demonstrate that LoRA can effectively tune Stable Diffusion to produce images that challenge the CNN model's ability to distinguish real from AI-generated images. The significant drop in accuracy suggests that LoRA introduces modifications that reduce detectable "fingerprinting," increasing the photorealistic quality of generated images. These findings emphasize the importance of incorporating LoRA-tuned datasets into classifier training to enhance robustness against advanced AI image generation techniques.

\vspace{-2ex}
\paragraph{Using DenseNets.}
The DenseNet model demonstrated robust performance, outperforming the CNN model across all evaluation scenarios. Table~\ref{table:densenet_cnn_accuracy_part1} and ~\ref{table:densenet_cnn_accuracy_part1} combine results for DenseNet and CNN models trained and tested on various Stable Diffusion datasets. DenseNet consistently achieved higher accuracy compared to the CNN model, both for real and AI-generated images.

DenseNet's performance was particularly notable for blurred images, a challenging scenario where patterns indicative of Stable Diffusion may be obscured. When tested on the CIFAKE-SD2.1-Blurred dataset, DenseNet demonstrated a significant advantage over CNN, achieving a 15.75\% higher overall accuracy and a 23.29\% higher accuracy in detecting AI-generated images. This resilience to Gaussian blurring underscores DenseNet's capacity to maintain strong performance even under conditions where image clarity is compromised.

DenseNet's architecture, with its dense connectivity and efficient feature reuse, consistently outperformed traditional CNN models across all scenarios. Its ability to excel on blurred datasets highlights its potential as a robust model for detecting AI-generated images, even in challenging real-world conditions.

\section{Conclusion}
The application of Convolutional Neural Networks (CNNs) has proven effective in distinguishing images generated by Stable Diffusion from authentic photographs. Notably, the performance of CNN-based detectors remains largely invariant to variations in image size, input prompts, and negative prompts provided to the Stable Diffusion model. However, these detectors exhibit significant sensitivity to the specific version of Stable Diffusion employed. Additionally, adversarial techniques, such as the application of Gaussian blurring or the use of Low-Rank Adaptation (LoRA), pose challenges to the accurate detection of AI-generated images.

To mitigate these vulnerabilities, the adoption of DenseNet architectures has shown promise. DenseNet demonstrates improved robustness against Gaussian blurring and achieves superior overall performance compared to CNNs in the detection of Stable Diffusion-generated images. This highlights its potential as a more reliable framework for addressing adversarial modifications in AI-generated content.

{\small
\bibliographystyle{ieee}
\bibliography{citations}}

\maketitle

\end{document}